\documentclass{bmvc2k}

\usepackage{float}
\usepackage{amsmath, amssymb}
\usepackage{graphicx}
\usepackage{hyperref}
\usepackage{url}
\usepackage{multirow}
\usepackage{booktabs}
\usepackage{longtable}
\usepackage{array}
\usepackage{makecell}

\title{ST-GDance: Long-Term and Collision-Free Group Choreography from Music}  %

\addauthor{Jing Xu}{jing.xu1@monash.edu}{1}
\addauthor{Weiqiang Wang\textsuperscript{1,\dag}}{weiqiang.wang@monash.edu}{}
\addauthor{Cunjian Chen\textsuperscript{1,\dag}}{cunjian.chen@monash.edu}{}
\addauthor{Jun Liu}{j.liu81@lancaster.ac.uk}{2}
\addauthor{Qiuhong Ke}{qiuhong.ke@monash.edu}{1}

\addinstitution{
 Department of Data Science and AI\\
 Monash University\\
 Melbourne, Australia
}
\addinstitution{
 School of Computing and Communications\\
 Lancaster University\\
 Lancaster, UK
}

\runninghead{XU, WANG, CHEN, LIU, KE}{ST-GDANCE}

\begin{document}

\maketitle

\begin{abstract}
Group dance generation from music has broad applications in film, gaming, and animation production. However, it requires synchronizing multiple dancers while maintaining spatial coordination. As the number of dancers and sequence length increase, this task faces higher computational complexity and a greater risk of motion collisions. Existing methods often struggle to model dense spatial-temporal interactions, leading to scalability issues and multi-dancer collisions. To address these challenges, we propose ST-GDance, a novel framework that decouples spatial and temporal dependencies to optimize long-term and collision-free group choreography. We employ lightweight graph convolutions for distance-aware spatial modeling and accelerated sparse attention for efficient temporal modeling. This design significantly reduces computational costs while ensuring smooth and collision-free interactions. Experiments on the AIOZ-GDance dataset demonstrate that ST-GDance outperforms state-of-the-art baselines, particularly in generating long and coherent group dance sequences. Project page: \url{https://yilliajing.github.io/ST-GDance-Website/}.

\end{abstract}

\section{Introduction}  
\label{sec:intro}
Dance is a highly expressive art form that uses body movements to convey emotion. With the rise of generative AI, music-to-dance generation has found increasing applications in film, cultural productions, gaming, and beyond~\cite{lee2021creators, valle2021transflower, yao2023dance}, assisting artists in crafting immersive experiences for both dancers and audiences. To engage the audience, dance must align with the rhythm and emotional tone of the music. Compared to single dance, group dance offers richer visual impact through spatial coordination, synchronized actions, and dynamic formation changes, making choreography significantly more complex. While existing works~\cite{le2023controllable, dai2025harmonious, yang2024codancers} and datasets~\cite{tsuchida2019aist, aiozGdance, zhang2025motion} have advanced the field, two major challenges remain.

\textbf{Generation of long sequences.} The goal of group dance generation is to synthesize complete and coherent motion sequences aligned with extended music compositions. Short sequences often fail to capture the structure and emotion of the full composition, such as such as the gradual rise before a beat drop in electronic dance music. Lodge~\cite{li2024lodge} addresses long-sequence generation via a two-stage method for solo dance, first generating key poses and then interpolating intermediate frames. Although this strategy improves scalability, it introduces several limitations: (a) errors can propagate between stages and (b) it relies on predefined motion patterns that cannot cover all dance variations, and the looping process slows down inference. More critically, it lacks global control, which is essential for group dance where dancers must stay aligned and move consistently. While end-to-end models could better capture global interactions, most group dance methods, such as GCD~\cite{le2023controllable} and TCDiff~\cite{dai2025harmonious}, use diffusion transformers. These models struggle with long sequences as their computational complexity scales quadratically with $N^2 \times L^2$ ($N$ and $L$ represent the size of the group and sequence length, respectively)~\cite{zhou2021informer, bulatov2024beyond}, resulting in inefficient performance and high resource demands. Efficiently generating high-quality, long group dance sequences thus remains an open challenge.

\textbf{Multi-dancer interaction modeling.} As previously mentioned, the artistic merit of group dance performances derives from precise inter-dancer coordination. Current choreography methods~\cite{le2023controllable, aiozGdance, tseng2023edge, yang2024codancers} typically represent group dance sequences by concatenating high-dimensional motion features and 3D coordinates across all performers, then processing this combined representation as a unified input vector. This approach discourages the model from effectively learning positional interactions, as it fails to preserve the independence and spatial correlations among individual dancer sequences. GCD~\cite{le2023controllable} addresses inter-dancer relationships using a simple cross-attention mechanism, which captures only vague spatial-temporal correlations. As a result, the generated group dance sequences often exhibit excessive motion uniformity. TCDiff~\cite{dai2025harmonious} also identifies this issue, referring to it as dancer ambiguity. To mitigate it, it introduces a trajectory-controllable diffusion model that explicitly extracts trajectory information. However, during trajectory generation, the model fails to account for spatial correlations among dancers’ trajectories, resulting in insufficient spatial constraints. This lack of spatial awareness often leads to unnatural overlaps or collisions between dancers during generation. Moreover, the dance-beat navigator module introduces computational overhead, rendering the model challenging to apply in long-term generation scenarios.

In summary, we identify that the efficiency challenges in long-term group dance generation mainly stem from two factors: the quadratic complexity growth in transformers with respect to temporal sequence length ($L^2$) and the number of dancers ($N^2$). Most existing methods treat the group as a single concatenated sequence of $N \times L$ tokens, processing each dancer-frame independently without explicitly modeling inter-dancer spatial relationships. This limits their ability to capture spatial correlations, leading to potential dancer ambiguity and movement collisions. To address these challenges, we propose \textbf{ST-GDance}, a highly efficient spatial-temporal model for long-term group dance generation. We decouple the generation task into spatial and temporal components. This allows us to replace the bulky attention module in DiT with: (a) a lightweight graph convolutional network (GCN)\cite{kipf2016semi}, which is well-suited for incorporating spatial awareness constraints for spatial modeling; and (b) adopting speed-up sparse attention techniques (local dependency transformer\cite{xie2024sana} and differential attention~\cite{ye2025differential}) for temporal modeling. Consequently, we attain lower computational costs with substantially fewer parameters when generating extended sequences.

\begin{figure}[H]
    \centering
    \bmvaHangBox{\includegraphics[scale=0.4]{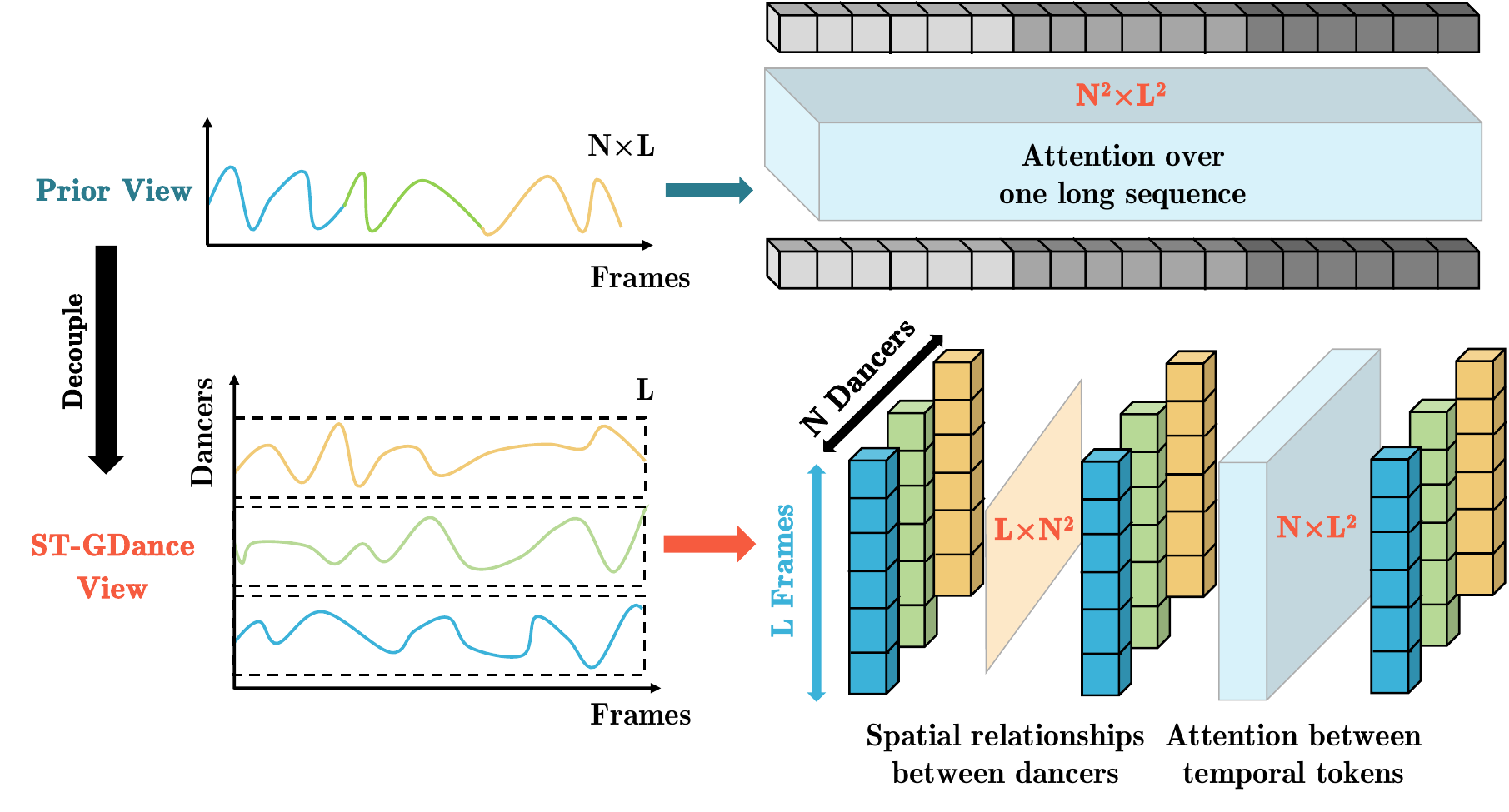}}
    \caption{Illustration of the overall framework for the proposed ST-GDance. We decouple the processing of long $N \times L$ sequences into spatial and temporal components to reduce computational complexity from $O\left( (NL)^2 \right)$ to $O(LN^2) + O(NL^2)$. Curves with different colors on the left represent data from different dancers. The overall complexity can be reduced to approximately $O(LN^2) + O(NL)$ (as will be discussed in Section~3.6).}
    \label{fig:design}  
\end{figure}

\begin{itemize}
    \item We propose ST-GDance, a novel spatial-temporal framework that decouples group dance generation into spatial and temporal components (Figure~\ref{fig:design}), enabling lightweight and modular processing.

    \item We propose a spatial modeling method that incorporates spatial distance-aware constraints, promoting structured and coordinated group choreography and preventing dance movement collisions.
    
    \item Our method avoids the traditional long-sequence paradigm and reduces the denoising complexity from $O\left( (NL)^2 \right)$ to nearly $O(L N^2) + O(NL)$, allowing efficient generation under limited resources, particularly for long frame sequences.
    
    \item Experiments on the AIOZ-GDance dataset show that our method outperforms existing approaches, especially in generating long group dance sequences.
\end{itemize}

\vspace{-2.5em} 
\section{Related Works}  
\textbf{Music-driven Single Dance Generation.} Music-to-dance generation aims to produce natural and coherent motion sequences~\cite{joshi2021extensive}. Research has evolved from traditional methods~\cite{kovar2002motion,kim2003rhythmic,safonova2007construction} to deep learning, with recent trends focusing on generative models~\cite{zhu2023human,yin2023dance,sun2020deepdance,ferreira2021learning}. Most approaches depend on music and motion history as inputs, which can lead to error accumulation and restrict diversity. Many also fail to model long-range dependencies, weakening coherence and novelty. With the availability of large datasets~\cite{lee2019dancing,li2021ai,li2023finedance}, generative models have become the dominant approaches. Auto-regressive models~\cite{alemi2017groovenet,yalta2019weakly} predict step-by-step but often drift. Multimodal methods like GANs~\cite{lee2019dancing}, VAEs~\cite{hong2022avatarclip}, and diffusion models~\cite{tseng2023edge} enhance diversity and fidelity. However, two key challenges remain: (a) most models are trained on short clips, limiting generalization to long sequences; and (b) single-dancer models struggle with group scenarios that require temporal alignment and spatial coordination.

\noindent\textbf{Spatial-temporal Framework in Motion Generation.}  
Spatial-temporal modeling is essential for motion tasks, capturing both spatial relations and temporal dynamics. In motion prediction, Transformer-based~\cite{aksan2021spatio} and graph-based~\cite{yin2021graph} models effectively capture joint dependencies and transitions. Video generation also benefits from spatial-temporal structures for coherent frame progression~\cite{zhu2023human}. Group dance is inherently spatial-temporal~\cite{zhang2025motion}, combining skeletal formations with dancer positioning. Graphs naturally represent spatial layouts, while temporal dynamics—driven by music and choreography—require long-range modeling. We thus adopt a spatial-temporal framework to jointly model formations and motion evolutions, improving consistency and simplifying long-term generation.

\noindent\textbf{Group Dance Generation.} Group dance generation has received increasing attention, with efforts focusing on coordinated multi-person motion~\cite{aliakbarian2020stochastic,arikan2002interactive,heusel2017gans}. However, current methods still face two core issues. First, earlier approaches~\cite{aliakbarian2020stochastic, heusel2017gans} often rely on convolutional or recurrent architectures to implicitly capture spatial relationships, which can result in unrealistic formations and occasional collisions between dancers. Second, Transformer-based models like GDanceR~\cite{le2023music}, GCD~\cite{le2023controllable}, and TCDiff~\cite{dai2025harmonious} suffer from high computational cost and performance degradation due to quadratic complexity in long sequences. To overcome these issues, we introduce a spatial-temporal decoupling approach that separates spatial and temporal modeling. This design improves computational efficiency and promotes spatial coordination and temporal consistency, resulting in more realistic and stable group performances.

\vspace{-1em} 
\section{Proposed Method}
\subsection{Preliminary}
Given a music sequence $\mathcal{M} = \{ m^i \}_{i=1}^{L}$, the goal is to generate a corresponding group dance sequence $\mathcal{X} = \{ x^i \}_{i=1}^{L}$, where $x^i = \left\{ x^{i,n} \right\}_{n=1}^{N}$, $L$ denotes the sequence length, and $N$ is the number of dancers. Diffusion models have shown strong capabilities in high-dimensional generative tasks such as images~\cite{saharia2022photorealistic}, audio~\cite{guo2024audio}, and videos~\cite{qi2023diffdance}. Following DDPM~\cite{ho2020denoising} and EDGE~\cite{tseng2023edge}, the forward and reverse diffusion processes are defined as:

\begin{equation}
q(x_t | x_{t-1}) = \mathcal{N}(x_t; \sqrt{1 - \beta_t}x_{t-1}, \beta_tI),
\end{equation}

\begin{equation}
p(x_{t-1} | x_t) = \mathcal{N}(x_{t-1}; \frac{1}{\sqrt{1 - \beta_t}}(x_t - \beta_t\mu_t), \sigma_t^2 I),
\end{equation}

\noindent where \( x_t \) is the noisy sample at time step \( t \), \( \beta_t \) controls the noise schedule, \( \mu_t \) is the model-predicted mean, and $\sigma_t^2$ is the variance of the reverse process noise. The forward process adds noise step-by-step, while the reverse process learns to recover clean data from noise. 

\vspace{-1em} 
\subsection{Overview of the Method}
\begin{figure}[H]
    \centering
    \bmvaHangBox{\includegraphics[width=0.9\linewidth]{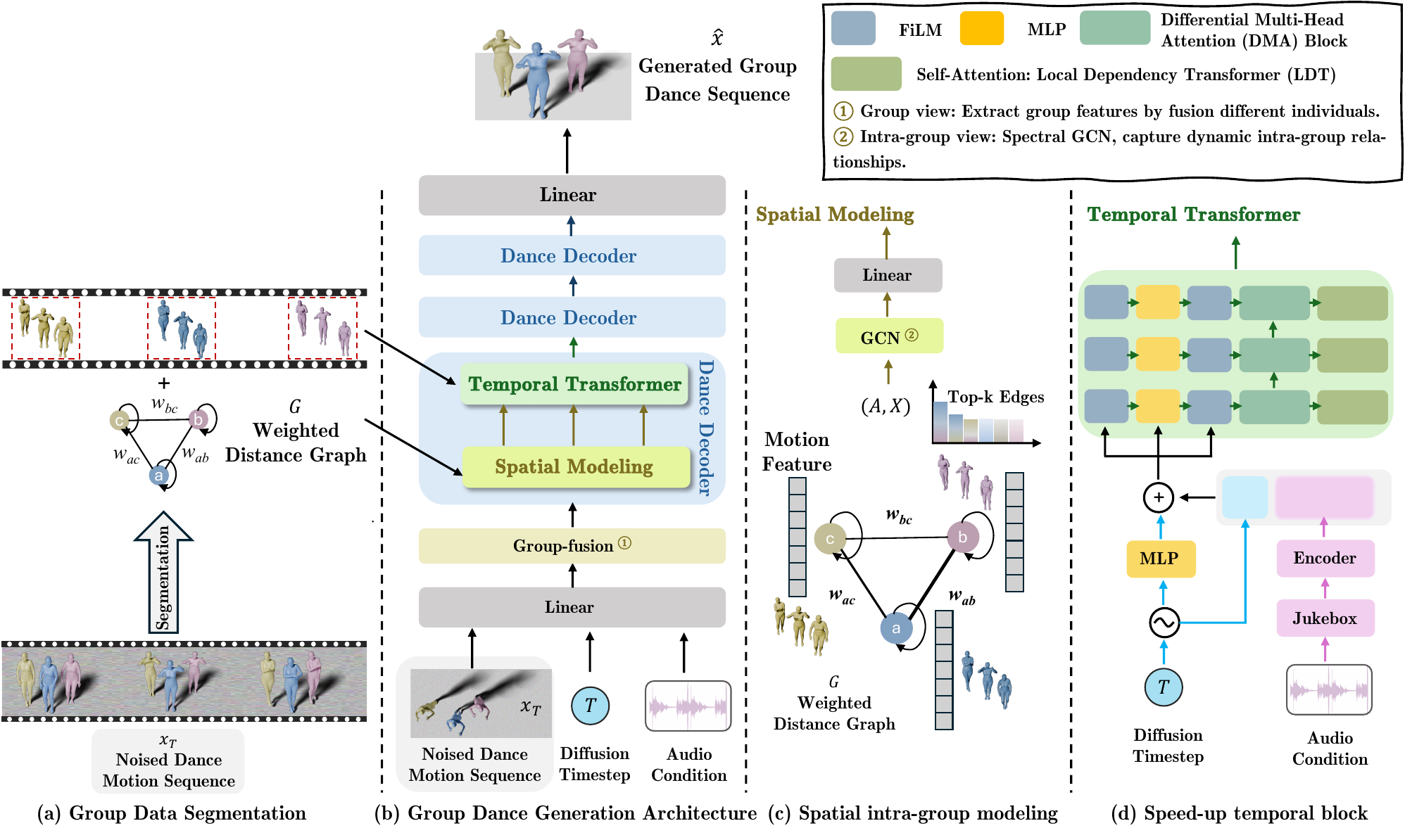}}
    \caption{The proposed framework for spatial-temporal group dance generation.}
\label{fig:framework}
\end{figure}

To generate realistic dance sequences, we use a spatial-temporal dance decoder during the denoising process as shown in Figure~\ref{fig:framework}. Given input $X_t$, we first apply a \textbf{Group Fusion block}~\cite{dai2025harmonious} which treats different dancers as a group to reduce dancer ambiguity with minimal memory usage: 

\vspace{-0.5em} 
\begin{equation}
\mathbf{X}_t^{\text{group}} = \text{GroupFusion}(\mathbf{X}_t).
\end{equation}

\noindent To enhance spatial interactions between dancers, we apply a distance-based \textbf{graph convolutional network (GCN)} with the calculated distance graph $\mathcal{G}$:

\begin{equation}
\mathbf{X}_t^{N} = \text{GCN}(\mathbf{X}_t^{\text{group}}, \mathcal{G}).
\end{equation}

\noindent For temporal coherence, we integrate a \textbf{Transformer-based decoder} in which we apply two speed-up attention techniques to model per-dancer frame-wise correlations:

\begin{equation}
\mathbf{X}_t^{i} = \text{TemporalTransformer}(\mathbf{X}_t^{i}, \mathbf{M}_t^{i}), \quad i \in [1, N].
\end{equation}

\vspace{-2em} 

\subsection{Spatial Distance-aware Graph Convolutional Network}
In group dance modeling, each frame involves two key aspects: interactions among dancers and overall group structure. Many prior works neglect the former. To model intra-group spatial relationships, we adopt a fully connected weighted graph, where each dancer is a node and edge weights are based on pairwise Euclidean distances, as illustrated in the group interaction module of our framework (see Figure~\ref{fig:framework} (d)). Let $N$ be the number of individuals, the adjacency matrix $A \in \mathbb{R}^{N \times N}$ for graph $G$ is defined as:

\begin{equation}
A_{ij} = \frac{1}{| p_i - p_j | + \epsilon},
\end{equation}

\noindent where $p_i$ and $p_j$ are the 2D ($x$, $y$) coordinates of individuals $i$ and $j$, and $\epsilon$ prevents division by zero while also serving as a spatial constraint to keep $A_{ij}$ non-zero, avoiding collisions and dancer ambiguity.

Given an input sequence $\mathbf{X} \in \mathbb{R}^{N \times L \times d}$, where $L$ is the sequence length and $d$ is the feature dimension, GCN propagates features as:

\begin{equation}
H^{(l+1)} = ReLU \left( \tilde{\mathbf{A}} H^{(l)} W^{(l)} \right),
\end{equation}

\noindent where $\tilde{\mathbf{A}} = D^{-1/2} \mathbf{A} D^{-1/2}$ is the normalized adjacency matrix, and $W^{(l)}$ is a learnable transformation. To reduce computation, we apply \textbf{top-k edge pruning}, retaining only the $k$ most significant edges, yielding per-layer complexity:

\begin{equation}
O_{\text{GCN}} = O(L N^2), 
\end{equation}

By incorporating a lightweight GCN to model spatial correlations, we significantly reduce redundant cross-attention in Transformers. This not only improves efficiency and reduces complexity, but also enables finer spatial control and helps avoid motion collisions.

\subsection{Temporal Dependency Modeling}
Following the design of EDGE~\cite{tseng2023edge}, we first incorporate timestep information through token concatenation with music conditioning and modulate features using FiLM (Feature-wise Linear Modulation) blocks~\cite{perez2018film} as shown in Figure~\ref{fig:framework} (c).

To further model temporal dependencies, we replace standard multi-head Attention with Differential Attention from DIFF Transformer~\cite{ye2025differential}, which enhances information filtering by computing attention as the difference between two sets of attention heads. This module is depicted in the temporal decoder of our framework (see Figure~\ref{fig:framework} (c)). Given one person $X \in \mathbb{R}^{L \times d}$, we project it into $Q_1, Q_2, K_1, K_2 \in \mathbb{R}^{L \times d}$ and $V \in \mathbb{R}^{L \times 2d}$, where $Q$, $K$ are obtained by splitting the last dimension of linear projections $X W^Q$ and $X W^K \in \mathbb{R}^{L \times 2d}$. 

\vspace{-0.5em} 

\begin{equation}
[Q_1; Q_2] = X W_Q, \quad [K_1; K_2] = X W_K, \quad V = X W_V, 
\end{equation}

\vspace{-0.7em} 
\begin{equation}
A_1 = \text{softmax} \left( \frac{Q_1 K_1^T}{\sqrt{d}} \right), \quad
A_2 = \text{softmax} \left( \frac{Q_2 K_2^T}{\sqrt{d}} \right),
\end{equation}

\vspace{-0.7em}  
\begin{equation}
\text{DiffAttn}(X) = (A_1 - \lambda A_2) V,
\end{equation}

\vspace{-0.3em}  
\noindent where $\lambda$ is a learnable suppression coefficient. Although the raw attention score computation involves $O(L^2)$ complexity, DiffAttn allows shared projections and attention filtering, resulting in near-linear complexity $O(L)$ in practice~\cite{ye2025differential}  (also see Table~\ref{tab-abla-efficiency}). Thus, the final complexity for a group of $N$ dancers would be $O(NL)$. 

To efficiently model long-range temporal dependencies, we adopt the Local Dependency Transformer (LDT) from SANA~\cite{xie2024sana} instead of traditional self-attention. Unlike standard attention with $O(L^2)$ complexity, LDT restricts computation to a local window $W \ll L$, greatly improving efficiency. Given input $X \in \mathbb{R}^{L \times d}$, LDT computes:   

\vspace{-0.5em}  


\begin{equation}
\mathrm{LDT\text{-}Attention}(Q, K, V) = 
\frac{
\mathrm{ReLU}(Q) \cdot \left( \sum_{j=1}^{N} \mathrm{ReLU}(K_j)^\top V_j \right)
}{
\mathrm{ReLU}(Q) \cdot \left( \sum_{j=1}^{N} \mathrm{ReLU}(K_j)^\top \right)
}
\quad 
\end{equation}

\vspace{-0.5em} 

Firstly, MLP and ReLU project $Q$, $K$, $V$ from $L \times d$ to $W \times d$, enabling efficient local attention. Then, the attention is reformulated by computing over $K$ and $V$ instead of $Q$ and $K$ first, greatly reducing the cost. The local window design not only strengthens temporal dependency modeling, leading to more precise and coherent dance generation, but also reduces the complexity to:

\vspace{-0.5em}  

\begin{equation}
O_{\text{LDT}} = O(L W) \approx O(L),
\end{equation}

\noindent where $W$ can be omitted as it is a constant much smaller than $L$ in our empirical settings. Therefore, the final complexity of a group of $N$ dancers over $L$ frames is $O(NL)$.

\vspace{-1em}  
\subsection{Loss Functions}
Following the loss design in EDGE~\cite{tseng2023edge}, we combine a basic reconstruction loss with three auxiliary terms, where \(L_{\text{joint}}\) enforces accurate joint positions, \(L_{\text{vel}}\) encourages temporal smoothness, and \(L_{\text{contact}}\) ensures realistic foot contact.

\vspace{-0.5em}  

\begin{equation}
Loss = L_{\text{simple}} + \lambda_{\text{pos}} L_{\text{pos}} + \lambda_{\text{vel}} L_{\text{vel}} + \lambda_{\text{contact}} L_{\text{contact}}.
\end{equation}

Here, $\lambda_{\text{pos}}$, $\lambda_{\text{vel}}$, and $\lambda_{\text{contact}}$ are hyper-parameters that weight the contributions of the position, velocity, and contact consistency losses, respectively. The specific values will be provided in the supplementary material. These components collectively enable the model to produce more realistic and physically consistent dance motions.

\vspace{-0.7em}
\subsection{Overall Complexity Analysis}
The dance decoder, shown in Figure~\ref{fig:framework}, combines a spatial GCN and a temporal Transformer with Differential Attention and Local Dependency Transformer applied on separate layers, leading to the following computational complexity:

\begin{equation}
\begin{aligned}
O_{\text{total}} &= O_{\text{GCN}} + O_{\text{LDT}} + O_{\text{Diff}} \\
                 &\approx O(L N^2) + O(NL).
\end{aligned}
\end{equation}

The key influencing factors are: (1) The number of individuals $N$, which affects the $O(L N^2)$ quadratic term from the GCN; (2) The sequence length $L$, where full attention would introduce $O(NL^2)$ quadratic complexity, but LDT reduces it to near $O(NL)$; and (3) While the theoretical complexity remains quadratic $O(NL^2)$, DiffAttn achieves near-linear empirical scaling $O(NL)$ through shared projections and attention filtering. By employing LDT and Differential Attention instead of standard multi-head attention, we achieve substantial complexity reduction from $O(NL^2)$ to $O(NL)$, making our approach significantly more efficient for large-scale motion sequences.

\vspace{-1.3em}  
\section{Experiment}
\subsection{Experimental Settings}
\textbf{Implementation Details.} We train our model on a dual RTX 4090 GPU setup. For the short-term task, the sequence length was \textbf{120 frames with 3 dancers (for group dance)}, and \textbf{400 frames with 3 dancers (for group dance)} for the long-term task. The model was trained for 2000 epochs using the Adam optimizer with a learning rate of 0.0002 and a dynamic scheduler. The loss function includes weighted terms for position, velocity, and contact constraints, with $\lambda_\mathrm{pos}$, $\lambda_\mathrm{vel}$, and $\lambda_\mathrm{contact}$ set differently for short- and long-term tasks to reflect motion dynamics.

\noindent\textbf{Dataset.} We use the AIOZ-GDance dataset~\cite{aiozGdance}, containing 16.7 hours of group dance videos with synchronized music and 3D motion data across 4,000 dancers, 7 dance styles, and 16 music genres. Dataset partitioning follows the protocol of~\cite{le2023music}. As the dataset consists of short clips segmented from longer sequences, we reconstruct full-length sequences for long-term tasks by grouping clips with the same sequence ID in the metadata.

\noindent\textbf{Baselines.} We compare ST-GDance against the following state-of-the-art group and single dance generation models, as well as methods for long-sequence tasks: (1) GCD~\cite{le2023controllable}: Combines group-based contrastive learning with diffusion for enhanced group motion representation. (2) CoDancers~\cite{yang2024codancers}: Ensures synchronized group dance generation via choreographic units aligned with music. (3) TCDiff~\cite{dai2025harmonious}: Integrates trajectory control into a diffusion framework to better model time-series dynamics. (4) Lodge~\cite{li2024lodge}: Designed for long-term single dance generation using a coarse-to-fine strategy guided by dance primitives. (5) EDGE~\cite{tseng2023edge}: Enables editable music-conditioned single dance generation with flexible control.

\noindent\textbf{Evaluation Metrics.} For evaluation, we follow~\cite{aiozGdance,yang2024codancers} using both single- and group-level metrics. Individual metrics include Frechet Inception Distance (FID) and Generation Diversity (Div); group-level metrics include Group Motion Realism (GMR), Group Motion Correlation (GMC), and Trajectory Intersection Frequency (TIF). Model efficiency is also measured via FLOPs, parameter count, and training / inference time.

\vspace{-1em}  

\subsection{Quantitative Results}

To demonstrate the effectiveness of our approach, we present a comparison with other state-of-the-art methods on short- and long-term in both group and single dance generation tasks in Table~\ref{tab-combined_all_grouped}. Since TCDiff was not open-sourced at the time of writing, its results are cited from the original paper under the same settings.

\noindent\textbf{Short-term Generation.} EDGE performs poorly on group dance metrics, especially GMR and TIF, as it was originally designed for single-person generation. It achieves strong FID in single dance, but with limited diversity. Our method achieves competitive performance in group dance with high efficiency and good coherence. For single dance, it consistently ranks among the top methods in FID and PFC, balancing accuracy and efficiency. Though TCDiff and GCD excel in group dance, they fall short in single-dancer cases. Overall, our model shows better stability and diversity in both tasks.

\vspace{-0.7em}  
\begin{table}[H]
\centering
\caption{Combined evaluation of generation quality and computational efficiency. \textbf{Best} and \underline{runner-up} scores are highlighted. ``Single Dance'' refers to a single person dancing for the same duration. ``Inf Time'' stands for the average time taken for model generation.}
\resizebox{\textwidth}{!}{
\begin{tabular}{c|c|ccc|ccc|c|c|c|c}
\specialrule{1pt}{0pt}{0pt}
\multirow{2}{*}{Setting} & \multirow{2}{*}{Method} 
& \multicolumn{3}{c|}{\textbf{Group Dance}} 
& \multicolumn{3}{c|}{\textbf{Single Dance}} 
& \multicolumn{4}{c}{\textbf{Efficiency}} \\
\cline{3-12}
& & GMR↓ & GMC↑ & TIF↓ & FID↓ & Div↑ & PFC↓ 
& \makecell{FLOPs\\(GFLOPs)} & \makecell{Params\\(M)} & \makecell{Train Time\\ / epoch \\(min)} & \makecell{Inf Time\\(s)} \\
\specialrule{1pt}{0pt}{0pt}

\multirow{5}{*}{\textbf{Short-term}}
& EDGE~\cite{tseng2023edge}         & 70.26 & 62.24 & 0.42 & 33.24 & 10.11 & 3.07 & \underline{12.50} & \underline{52.13} & \textbf{0:24} & 8.59 \\
& GCD~\cite{le2023controllable}     & 30.22 & 80.22 & 0.19 & 39.24 & 9.64 & 2.53 & 27.69 & 62.16 & 1:04 & \underline{4.75} \\
& CoDancers~\cite{yang2024codancers} & 26.34 & 74.22 & \textbf{0.10} & \textbf{23.98} & 9.48 & 3.26 & -- & -- & -- & -- \\
& TCDiff~\cite{dai2025harmonious}   & \underline{15.36} & \textbf{82.77} & \underline{0.11} & 37.31 & \textbf{14.01} & \textbf{0.51} & -- & -- & -- & -- \\
& \textbf{ST-GDance}                & \textbf{14.76} & \underline{81.24} & \underline{0.11} & \underline{28.87} & \underline{12.83} & \underline{0.97} & \textbf{6.75} & \textbf{50.21} & \underline{0:37} & \textbf{3.47} \\
\specialrule{1pt}{0pt}{0pt}

\multirow{5}{*}{\textbf{Long-term}}
& EDGE~\cite{tseng2023edge}         & 76.24 & 59.86 & 0.57 & 42.79 & 6.32 & 4.23 & 23.20 & 53.83 & 1:02 & 35.59 \\
& Lodge~\cite{li2024lodge}          & 67.32 & 62.31 & 0.41 & \textbf{30.19} & 9.31 & \underline{3.21} & \underline{13.23} & \underline{38.37} & \textbf{0:46} & \underline{25.93} \\
& GCD~\cite{le2023controllable}     & 42.64 & 67.85 & 0.20 & 32.42 & \underline{8.94} & 3.95 & 31.60 & 62.29 & 2:51 & 31.42 \\
& CoDancers~\cite{yang2024codancers} & \underline{40.73} & \underline{68.22} & \underline{0.14} & 31.92 & 8.45 & 4.03 & -- & -- & -- & -- \\ 
& \textbf{ST-GDance}                & \textbf{31.76} & \textbf{75.18} & \textbf{0.13} & \underline{30.21} & \textbf{11.35} & \textbf{3.11} & \textbf{8.12} & \textbf{50.35} & \underline{1:25} & \textbf{17.84} \\
\specialrule{1pt}{0pt}{0pt}
\end{tabular}
}
\label{tab-combined_all_grouped}
\end{table}

\vspace{-1.2em}  

\begin{table}[H]
\caption{Ablation study of ST-GDance. We remove different modules and evaluate on short-term group (3 dancers) and single (1 dancer) dance generation. The best-performing results are highlighted in bold.}
\centering
\resizebox{\textwidth}{!}{
\begin{tabular}{ccc|ccc|ccc|c|c|c|c}
\specialrule{1pt}{0pt}{0pt}
\multicolumn{3}{c|}{\textbf{Ablated Modules}} & 
\multicolumn{3}{c|}{\textbf{Group Dance}} & 
\multicolumn{3}{c|}{\textbf{Single Dance}} & 
\multicolumn{4}{c}{\textbf{Efficiency}} \\
\cline{1-13}
\makecell{SMB} & \makecell{LDT} & \makecell{DMA} 
& \makecell{GMR↓} & \makecell{GMC↑} & \makecell{TIF↓} 
& \makecell{FID↓} & \makecell{Div↑} & \makecell{PFC↓} 
& \makecell{FLOPs\\(GFLOPs)} & \makecell{Params\\(M)} & \makecell{Train Time\\(min)} & \makecell{Inf Time\\(s)} \\
\specialrule{1pt}{0pt}{0pt}

\multicolumn{1}{c|}{} & \multicolumn{1}{c|}{\checkmark} & \checkmark 
& \multicolumn{1}{c|}{28.58} & \multicolumn{1}{c|}{71.24} & 0.16 
& \multicolumn{1}{c|}{31.85} & \multicolumn{1}{c|}{12.30} & 1.13 
& 26.87 & 58.31 & 1:35 & 28.41 \\

\multicolumn{1}{c|}{\checkmark} & \multicolumn{1}{c|}{} & \checkmark 
& \multicolumn{1}{c|}{16.54} & \multicolumn{1}{c|}{79.89} & 0.12 
& \multicolumn{1}{c|}{26.74} & \multicolumn{1}{c|}{11.92} & 1.15 
& 13.73 & 51.46 & 1:48 & 24.31 \\

\multicolumn{1}{c|}{\checkmark} & \multicolumn{1}{c|}{\checkmark} &  
& \multicolumn{1}{c|}{19.76} & \multicolumn{1}{c|}{79.24} & 0.13 
& \multicolumn{1}{c|}{25.62} & \multicolumn{1}{c|}{11.74} & 1.35 
& 13.73 & 51.43 & \textbf{0:39} & \textbf{15.62} \\

\specialrule{1pt}{0pt}{0pt}
\multicolumn{3}{c|}{\textbf{ST-GDance (Full)}} 
& \multicolumn{1}{c|}{\textbf{14.76}} & \multicolumn{1}{c|}{\textbf{81.24}} & \textbf{0.11} 
& \multicolumn{1}{c|}{\textbf{28.87}} & \multicolumn{1}{c|}{\textbf{12.83}} & \textbf{0.97} 
& \textbf{8.12} & \textbf{50.35} & 1:25 & 17.84 \\
\specialrule{1pt}{0pt}{0pt}
\end{tabular}
}
\label{tab-abla-efficiency}
\end{table}

\vspace{-1em}  


\noindent\textbf{Long-term Generation.} ST-GDance consistently surpasses baselines across group and single dance. It achieves the best group dance results, lowering GMR, enhancing GMC, and reducing temporal inconsistency for smoother motion. In single dance, it outperforms others in diversity, including Lodge, while remaining competitive in FID. These results suggest that ST-GDance captures complex motion while maintaining coherence, improving diversity by $\sim$20\% over prior methods. Comparison results across different group sizes, obtained by retraining models for each setting, along with visualizations, will be provided in the supplementary material.

\noindent\textbf{Efficiency Analysis.} We assess the efficiency of ST-GDance against diffusion-based baselines like EDGE, Lodge (first stage) and GCD. Results are in the experimental setting of group dance. We also test variants using linear and differential attention. As shown in Table~\ref{tab-combined_all_grouped}, ST-GDance achieves strong performance with 8.12 GFLOPs and 50.35 M parameters. Its training and inference times (1:35 min and 17.84s) highlight its efficiency. The results in the three tables highlight the practical advantages of ST-GDance over heavier models such as GCD.

\noindent\textbf{Effectiveness of Different Blocks.} To evaluate the contribution of each component in ST-GDance, we conduct an ablation study on the spatial modeling block (SMB), local dependency transformer (LDT), and differential multi-head attention block (DMA) in the temporal transformer. By selectively removing these modules, we assess their individual impact on group and single dance generation. As shown in Table~\ref{tab-abla-efficiency}, removing the SMB module (first row) eliminate spatial structural constraints, resulting in the most significant degradation in both efficiency and generation quality—particularly in the group dance setting. For example, inference time extends to 28.41 seconds and GMR increases notably. In contrast, the second and third rows retain the temporal-spatial decoupling modeling provided by SMB, keeping the model complexity at a manageable $O(LN^2) + O(NL)$. The key difference lies in the temporal acceleration strategy: while the LDT module achieves better inference efficiency than the differential multi-head attention module, it also omits more information, leading to a slight drop in generation quality. Overall, the ST-GDance model achieves the best trade-off between performance and efficiency.

\vspace{-1em}  

\subsection{Qualitative Results}

\vspace{-1.3em}  
\begin{figure}[H]
    \bmvaHangBox{\includegraphics[width=0.95\linewidth]{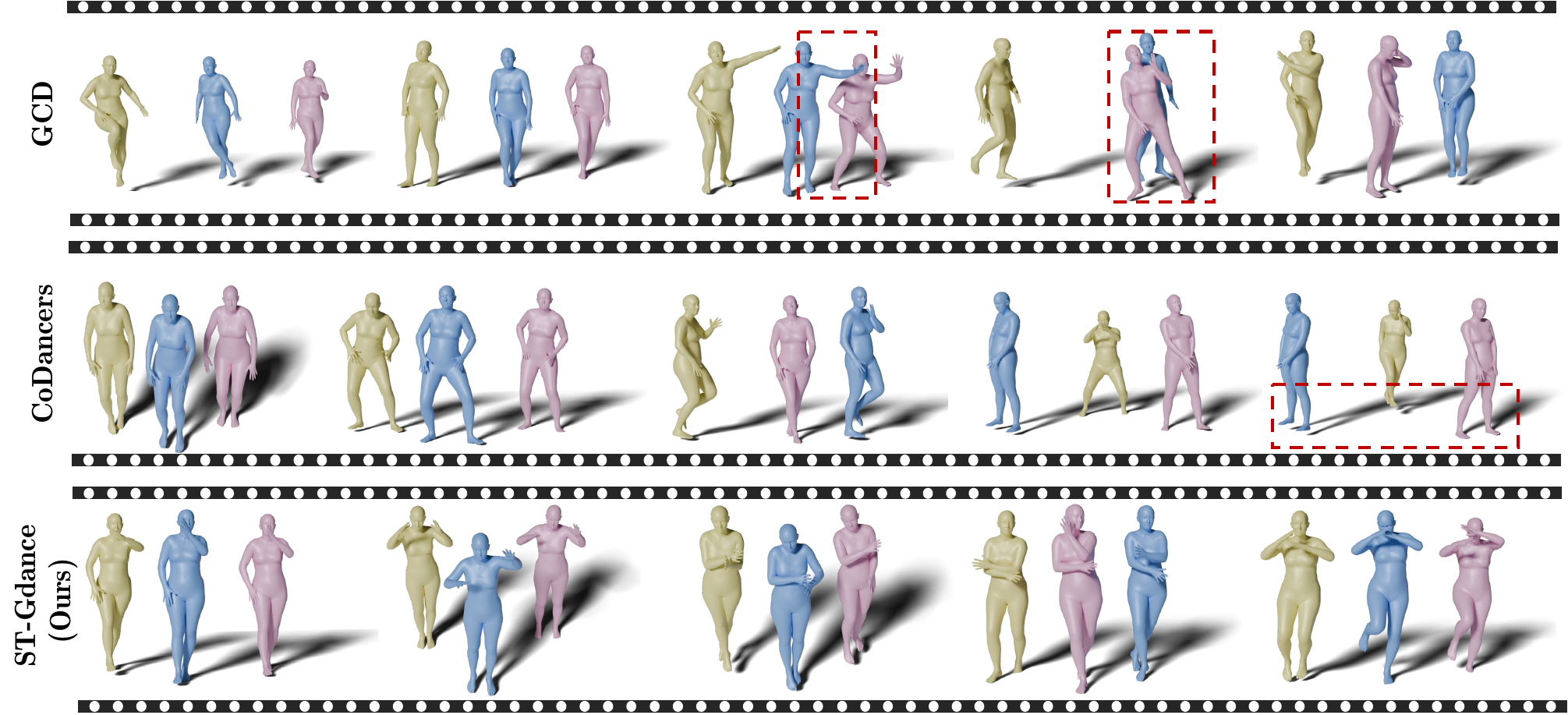}}
    \caption{Qualitative comparison of generated dance sequences across different models.} 
\label{fig:quali}
\end{figure}

Visual comparisons of dance sequences generated by our method, Co-Dancers, and GCD are shown in Figuree~\ref{fig:quali}.As highlighted by the red dashed boxes, GCD results exhibit spatial collisions or limb occlusions, affecting visual quality and coherence. CoDancers produces more harmonious motion but occasionally shows inconsistent foot placement relative to the ground, reducing realism. In contrast, our method produces smoother transitions and better spatial alignment, with fewer positional overlaps thanks to the spatial distance-aware constraints from GCN. These results indicate superior motion fluency and spatial consistency, leading to more natural and visually appealing sequences.

\section{Conclusion}
In conclusion, we propose ST-GDance, a novel spatial-temporal framework for efficient long-term and collision-free group dance generation. By combining lightweight graph convolutions with sparse attention, our approach overcomes the computational bottlenecks of transformer-based models, enabling scalable generation across many dancers and long sequences. Experiments on the GDance dataset show that ST-GDance outperforms state-of-the-art methods in both group and single dance tasks, significantly improving motion coherence and diversity. Its efficiency further supports real-time applications. We believe ST-GDance offers a promising direction for future research, with potential extensions to broader motion generation domains.

\section*{Acknowledgements}
This research was supported by the Australian Government through the Australian Research Council's DECRA funding scheme (Grant No.: DE250100030).

\end{document}